\newif\ifpeerreview   
\peerreviewtrue          
\peerreviewfalse         

\newif\ifpreprint   
\preprintfalse           
\preprinttrue            
\ifpeerreview
\documentclass[10pt,peerreview,a4paper,twocolumn] {IEEEtran}  
\else
\documentclass[10pt,conference,a4paper,twocolumn] {IEEEtran}  
\fi

\usepackage[switch]{lineno} 
\usepackage{graphicx}
\usepackage{epic}
\usepackage{figsize}
\usepackage{float}
\usepackage{hyperref}
\usepackage[section]{placeins}
\usepackage{cite}
\usepackage{amsmath}
\usepackage{amssymb}
\usepackage{amsfonts}
\usepackage{flushend} 
\usepackage{verbatim} 
\usepackage{enumitem} 
\usepackage{booktabs} 
\usepackage{xcolor}   
\newcommand{\email}[1]{{\fontfamily{cmtt}\selectfont#1\normalfont}}

\IEEEoverridecommandlockouts
\title{Selecting Datasets for Evaluating an Enhanced Deep Learning Framework\\
}

\author{\IEEEauthorblockN{Kudakwashe Dandajena${}^1$, Isabella M. Venter${}^2$, Mehrdad Ghaziasgar${}^2$, and Reg Dodds${}^2$} 
\IEEEauthorblockA{\emph{${}^{1,2}$Department of Computer Science, University of the Western Cape}\\
${}^1$\email{3986658@myuwc.ac.za}, 
${}^2$\email{\{iventer,mghaziasgar,rdodds\}@uwc.ac.za}\\
}}
\ifpreprint
\fi
\begin{document}
\ifpeerreview
\IEEEpeerreviewmaketitle
\else
\maketitle
\ifpreprint
\thispagestyle{plain}  
\fi
\fi
\begin{abstract}
A framework was developed to address limitations associated with existing 
techniques for analysing sequences. This work deals with the steps followed to 
select suitable datasets characterised by discrete irregular sequential 
patterns.  To identify, select, explore and evaluate which datasets from 
various sources extracted from more than 400 research articles, an interquartile 
range method for outlier calculation and a qualitative Billauer's algorithm 
was adapted to provide periodical peak detection in such datasets.
 The developed framework was then tested using the most appropriate datasets.
 The research concluded that the financial market-daily currency exchange 
domain is the most suitable kind of data set for the evaluation of the designed 
deep learning framework, as it provides high levels of discrete irregular 
patterns. 
\end{abstract}

\vspace{0.2cm}

\section{Introduction}
The current technological transformation, named the fourth industrial revolution 
(4IR)~\cite{schwab17} has positively transformed society by its: abundance 
of data; enhanced connectivity; industrial and workplace automation; autonomous 
and intelligent agents; artificial intelligence (AI) solutions; the Internet 
of things (IoT) technologies, etc.~\cite{lee18}. Every 
other industrial revolution, produced a different type of cutting-edge 
technology which had its own challenges. Institutions and individuals are 
investing in a considerable amount of resources such as capital, human talent,
 infrastructure, hardware platforms, environments and software tools to 
remain competitive and sustainable. It is important to note that research 
within 4IR, is assisting science to improve the lives of people. This 
can be validated by how researchers such as Geoffrey Hinton, Yoshua Bengio,
 Ian Goodfellow, Andrew Ng, etc., have leveraged developments 
for the advancement of AI~\cite{goodfellow16}.

The underlying challenges associated with existing deep learning frameworks 
for the analysis of discrete irregular patterned complex sequences were 
identified~\cite{dandajena20}.  Specific concerns raised were: performance 
robustness; transparency of the methodology; literature consistency; internal 
and external architectural design and configuration issues.  Inconsistencies 
and discord in the literature highlight some of the challenges associated 
with existing approaches to the analysis of irregular sequences and makes 
it possible to address them.  It was suggested that addressing these challenges 
might lead to a systematic, accurate, stable, explainable and repeatable 
deep learning framework for the analysis of these types of datasets.
The explainability of any system is determined 
by the ability to present and explain its underlying features in a way that 
can be understood~\cite{doshi-velez17}.  
The relationship between data and deep learning solutions has become 
important for the development of better artefacts for analysis~\cite{sarma19}. 
Current interest in innovative developments in the field of AI is often 
driven by large scale data.  Data is both a push and a pull factor for the 
development of AI technologies, which are producing and consuming an unprecedented 
abundance of data.  For example data is produced by financial markets, social 
media, astronomy, weather, traffic, surveillance, etc.~\cite{srivastava14}. 
There is an emerging trend that datasets are characterised by 
a high level of irregularities as a result of incompleteness, extreme randomised 
patterns and noise~\cite{ souma19}. Dealing with such massive datasets 
remains challenging.
\begin{figure}
\centering
\centering
\includegraphics[bb=20 10 622 440,scale=0.379]{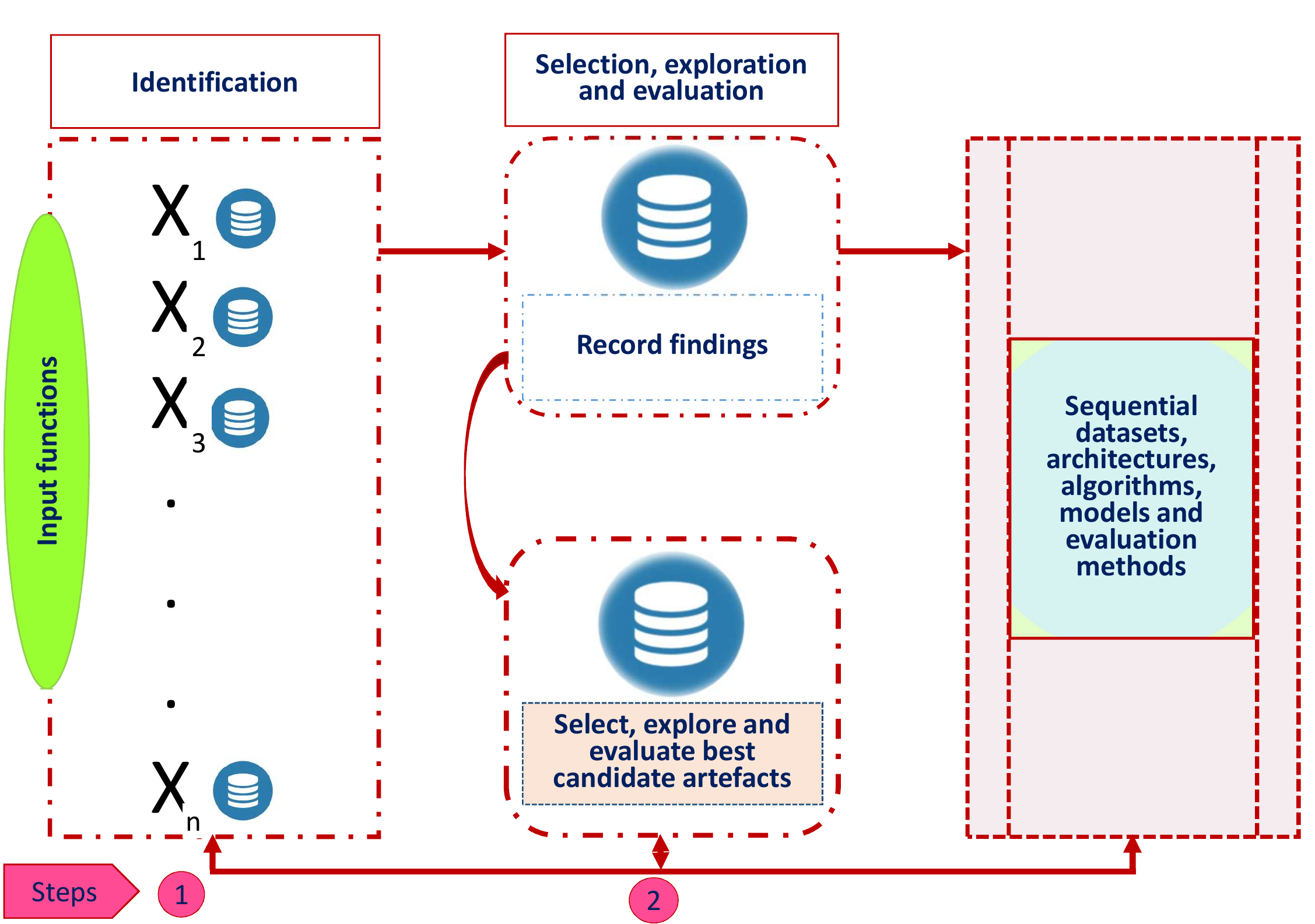}
%
\caption{SeLFISA framework dataset selection process}
\label{fig:datasetselection}
\end{figure}

This paper builds towards a framework named the \textit{Systematic enhanced deep 
Learning Framework for Irregular Sequential Analysis (SeLFISA)} for the analysis 
of these kinds of datasets. The framework will be a combination of architectures,
 algorithms and models aimed at predicting the future behaviour of complex 
sequences.  The intention is to produce an explainable and straightforward 
framework as part of a solution within the 4IR space. The initial step of 
such a framework involves selecting suitable and appropriate datasets, which 
is the focus of this paper.  The dataset selection process of the intended 
SeLFISA framework is shown in Fig.~\ref{fig:datasetselection}. As such,
 the aim of this research is to address the question: \textit{How should such a 
dataset be identified to evaluate the SeLFISA framework?} A design science 
research methodology guided the research.

Referring to the process in Fig.~\ref{fig:datasetselection}, a class of 
datasets with irregular patterns were identified as a key stage of developing 
better deep learning artefacts for sequential analysis. 
Seventy three sequential datasets from eight domains of different sources 
were extracted from 400 research articles, by means of 
a systematic literature review process~\cite{dandajena20}. 
Datasets, characterised with the most irregular patterns, were then identified 
for the evaluation of the deep learning framework. The daily financial market 
currency exchange domain provided high levels of discrete irregular patterned 
environments and the Pound to Dollar exchange rate was used for the evaluation.

\section{Materials and Methods}


The implementation process of the SeLFISA framework illustrated in Fig. 1 
involves detailed activities at each step to create a systematic, and 
repeatable way of analysing irregular discrete patterns in sequential environments.

The two steps in the framework shown in Fig. 1 were:

\noindent\textbf{Step 1}---Identify existing sequential datasets, analysis artefacts,
 and an evaluation mechanism using the systematic literature review process 
(SLR):  This stage creates an initial node that receives input from the SLR 
guided by reporting items for systematic reviews and meta-analyses 
(PRISMA)~\cite{moher09}.  Artefacts in the form of algorithms and models 
provide hints to guide the choice of implementation approaches, datasets 
and performance evaluation techniques. This stage focuses on trying 
to find initial answers to the following questions:
\begin{enumerate}
\item \textit{Which datasets are sequential in nature with irregular characteristics?}
\item \textit{Which artefacts in the form of algorithms and models have been 
      applied elsewhere to analyse such datasets?}
\item \textit{How were those artefacts evaluated in terms of determining their 
      performance?} 
\end{enumerate}
This step also creates a precondition for implementing the SeLFISA 
framework by ensuring that it produces repeatable and reliable output.

Failure to satisfy the initial condition will not only affect the entire 
development process of better artefacts to analyse such datasets but will 
negatively affect substantiation of the results. 

\textbf{Step 2}---Select, explore and evaluate candidate requirements:  Here 
identified datasets, classical algorithms for irregular sequential analysis, 
and evaluation mechanisms were loaded into a database of records.  
This was done to implement an ecosystem comprising high performance computing 
resources coupled with specific software libraries and tools.  An analysis 
that considered different dimensions had to be  made and applied to add 
insights  into eradicating elements of inconsistency and ambiguity that 
may affect future implementation steps and procedures. Once this analysis 
was done, a selection process commenced based on the following: 
\begin{enumerate}[label=\arabic*)]
\item \textbf{Datasets}---Select a dataset with most irregular patterns.
 Determine the number of irregular patterns of datasets by combining both 
descriptive numerical and visualization approaches. Check the levels of 
irregularity of selected datasets by considering the following:
\begin{enumerate}
\item \textit{Box and whisker plot}: apply statistical descriptive analysis of irregular 
patterns on all datasets by considering the interquartile range (IQR) outlier 
calculation~\cite{myat20}.  This approach illustrates the minimum and maximum 
values of any dataset.  The first, second and third quartiles Q1, Q2 and 
Q3 of the data are shown in the in the box plot. The difference between 
the minimum and maximum provides the range of values within the dataset.
 Finally, the difference between Q3 and Q1 provides an inter-quartile range 
(IQR) given by Equation (\ref{equation:one}):
\begin{equation}
IQR = Q3 - Q1\label{equation:one}
\end{equation}
Outliers in irregular patterns can easily be detected as those data points 
that are either one and a half times IQR below Q1 or above Q3, i.e. :
\begin{equation}
\mbox{Below }= Q1 - 1.5 \times IQR
\end{equation}
\begin{equation}
\mbox{Above }= Q3 + 1.5 \times IQR
\end{equation}

\item \textit{Billauer's algorithm} is used to validate outcomes of the box and whisker 
plot results and to detect local maxima and minima in a signal~\cite{xiao18}.  
It provides a graphical visualization analysis of peaks to measure the 
degree of irregularity of the original data environment.  The algorithm 
is then customised to provide irregular-pattern peak period 
detection (IPPD) which detects discrete peak values by searching for values 
that are surrounded by lower or larger values for maxima and minima across 
the y-axis and corresponding x-axis.  A look-ahead value for determining 
the look-ahead distance for a potential peak needs to be a set as a specific 
value to provide a maximum number of discrete peaks.
\item \textit{Further exploratory data analytics} then needs to be conducted 
for insights into how variables of the chosen sequential 
dataset are connected to each other.  Data description, data pre-processing,
 data munching, data cleaning, and exploratory data analysis all are executed 
under this umbrella procedure to understand a dataset in detail.
\end{enumerate}
\item \textbf{Artefacts}---Candidate artefacts were then selected, based on their 
intrinsic design, application, context and veracity.  At this stage, attention 
has to be focused on topical issues associated with common, debatable and 
contradicting issues highlighted by different authors.  Attention was also 
paid to the search space through variable clipping. 
\item \textbf{Evaluation}---Evaluation allowed for the development of a 
narrow list of multidimensional performance criteria. This was done to find 
the best metrics for each criterion that captures the requirements 
of the sequential analysis challenge. The multidimensional criteria encompass 
a basket of factors covering complexity accuracy, efficiency, stability,
 straightforwardness, explainability and repeatability.
\end{enumerate}

\section{Results}
A summary of the results for these implementation steps is given below:

\noindent\textbf{Step 1}---The identification of existing sequential datasets,
 analysis artefacts, and evaluation criteria.  This step is an extension of 
the systematic literature review work~\cite{dandajena20} which reviewed 
over 400 research articles from year 2015 to 2020 and narrowed them to the 33 
most relevant articles.  A summary of identified sequential models, architecture,
 datasets and evaluation techniques is shown in Table I.
\begin{table}
\caption{A summary of identified sequential datasets, analysis artefacts, and evaluation mechanisms}
\begin{tabular}{p{67pt}p{73pt}p{79pt}} 
\toprule                               
\textbf{Sequential Models and Architectures} & \textbf{Sequential Datasets}                                 & \textbf{Evaluation Techniques}\\
\midrule
$\geq34$ architectures                       & $\geq73$ datasets from $8$ domains of different sources.     &  $6$ evaluation techniques\\
\midrule
\multicolumn{3}{p{219pt}}{See Section~\ref{sec:annexures} for \url{Annexure_1_Results.pdf} in GitHub.}\\
\bottomrule
\end{tabular}
\end{table}\\

\noindent\textbf{Step 2}---The selection, exploration and the evaluation of candidate requirements
\begin{enumerate}[label=\roman*]
\item \textbf{Datasets}---At this stage, all datasets identified were 
collected into a bank of datasets---see Section~\ref{sec:annexures}  
\url{Annexure_2-Datasets_Models} in GitHub.
A hybrid high-end computational processing environment was provided 
by the South African Centre for High Performance Computing which provides 
NVIDIA GeForce MX130 graphical processing units (GPUs) and random access memory 
ranging between 20--210 GB, 10 TB HDD; a CUDA toolkit for GPU deployment;
 Anaconda distribution software with Python and Jupyter Notebook, and the 
Keras, TensorFlow, Pandas and other libraries.
  
  This dual mathematical and visualisation approach identified a dataset of 
financial daily exchange data between the GB Pound and the US Dollar from 
1990 to 2016~\cite{chang19} with the highest number of irregular patterns,
 i.e. 639 from 6135 daily records. The financial daily exchange data between 
the Japanese Yen and the US Dollar from the same period has around 24 irregular 
patterns from 5000 daily records. 

  Applying Billauer's algorithm produced irregular patterns peak period detection 
(IPPD) visualisation analysis of the daily exchange data between the GB 
Pound and the US Dollar.  See Fig. 2.  The financial daily exchange rate 
data between the GB Pound and the US Dollar became our primary candidate 
dataset for the SeLFISA framework evaluation, and the daily exchange rate 
data between the Japanese Yen and the US Dollar became the elected validation 
dataset to test performance stability and consistency.
\begin{figure}
\includegraphics[bb=10 310 810 575,scale=0.300]{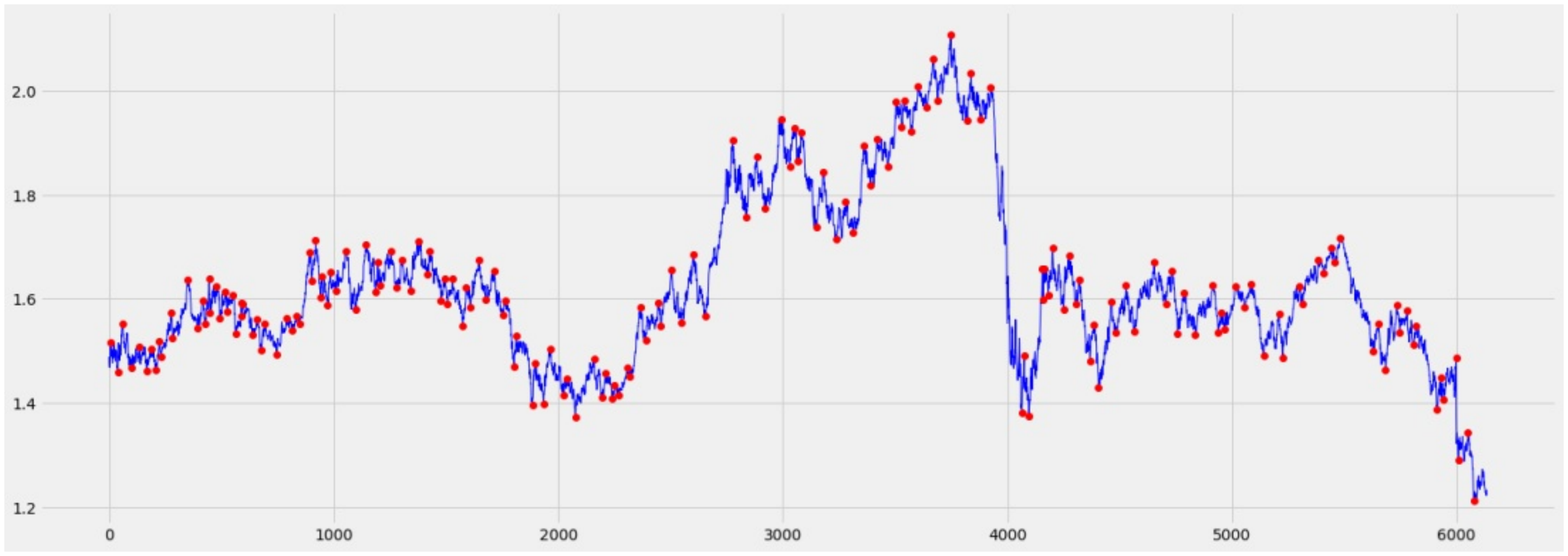}
\caption{IPPD visualisation analysis of daily exchange data between the 
GB Pound and the US Dollar from 1990 to 2016~\cite{chang19}}
\end{figure}
\item \textbf{Algorithms}---The experimental process identified 12 implementable 
or executable algorithms for sequential analysis from different authors 
which became the candidate artefacts. The selection process considered a 
combination of architecture uniqueness and referenced performance properties.
 Some of these algorithms were based on gated recurrent neural networks 
(RNN), autoencoders, convolutional neural networks, bidirectional mechanisms,
 attention mechanisms, ensemble techniques, deep and vanilla architectures.

    Specific architectural design features of these 12 selected algorithms were:
the gated LSTM architecture suggested~\cite{azlan19},~\cite{glenski19} 
and~\cite{chalvatzisa19}; the bidirectional mechanism combined with 
both LSTMs and GRUs influenced~\cite{sardelicha18}; the attention mechanism 
combined with gated neural networks~\cite{huang19}; deep convolutional 
neural network (CNN) ensemble with LSTM and an attention mechanism~\cite{makinen18}; 
a GRU~\cite{qin19} autoencoders combined 
with LSTM~\cite{wei19} and finally a deep gated recurrent 
neural network architectures made up of both GRU and LSTM~\cite{bai19}.

\item \textbf{Evaluation}---The following factors were considered as potential 
performance evaluation criteria with specific metrics: complexity measure 
through the total number of built parameters of every architecture, accuracy 
considered a \textit{mean absolute error} (MAE) which is robust in environments associated 
with discrete irregular patterns when measuring the average magnitude of 
the errors in a set of predictions, without considering their direction.
 \textit{Mean squared error} (MSE) provided an accuracy performance measure of the 
variance of the residuals and a quadratic scoring rule of \textit{root mean squared 
error} (RMSE) which gives a relatively high weight to large errors. The $R^2$ 
measures how well the independent variables in the linear regression model 
predict the dependent variable.  The lower the value of MAE, MSE and RMSE 
the more favourable the performance accuracy of the implemented model. Higher 
$R^2$ values indicate better performance.

Efficiency was measured as a ratio of execution time and the total number 
of parameters---whilst explainability and repeatability were identified qualitatively. 
\end{enumerate}

\section{Discussion}

There is still not much agreement on how to solve the challenges posed by 
the analysis of irregular sequential datasets~\cite{dandajena20}.  Contrary to existing 
frameworks---which are skewed towards probabilistic randomization or ad-hoc design 
approaches, which are prone to accuracy, stability, explainability and repeatability 
deficiencies---the SeLFISA framework aims at addressing these deficiencies~\cite{kuleshov18}.
To test the developed framework, appropriate datasets had to be identified.
  
\subsection{Datasets}
This paper has shown that analysing discrete irregular patterns 
or behaviour in sequential environments is more centred on a framework approach 
beyond just an architecture- or an algorithm- or a model-approach. The framework 
provides guidance for developing a robust analysis artefact. The suggested 
SeLFISA framework has proved to be useful---as it integrates outputs and outcomes 
from a wide range of research approaches to provide a systematic way to 
address performance, robustness, literature inconsistencies, straightforwardness 
and design limitations associated with existing sequential analysis techniques.
 Iterative steps of the SeLFISA framework deliver an explainable 
and understandable way of determining the levels of irregularity of such 
datasets. It applied Billauer's algorithm and mathematical and interquartile 
range outlier calculations to select, explore and evaluate a variety 
of sequential datasets~\cite{xiao18}. The daily exchange rate data between the GB Pound 
and the US Dollar has been identified as a dataset which is very suitable 
for learning because of high irregularity~\cite{chang19}. The daily exchange between the 
Japanese Yen and the US Dollar data can then be used as a validation dataset. 

\subsection{Algorithms}
On the other hand, models and algorithms designed through gated sequential 
architectures in the form of LSTMs and GRUs have been widely used in such 
analysis environments~\cite{azlan19},~\cite{glenski19},~\cite{chalvatzisa19}. 
Thus the guidance from the SeLFISA framework will influence the development of 
deep learning with artefacts that may demonstrate better performance over these 
suggested gated models.

\subsection{Evaluation}
This framework is useful when evaluating performance criteria with specific 
metrics for a particular analysis artefact. SelFISA provides a multidimensional 
perspective for examining critical design aspects and properties of chosen artefacts.

\section{Conclusion and Future Recommendations}
 The SelFISA framework was developed to address limitations associated with 
existing sequential analysis techniques. The identification, selection,
exploration and evaluation of datasets characterised by irregular discrete 
sequential characteristics using the SeLFISA framework provides a starting 
point towards the design of better performing, straightforward, explainable 
and understandable deep learning analysis artefacts. It creates a consistent 
shareable technical and literature platform which consists of a knowledge 
bank of implemented irregular sequential analysis frameworks, datasets, 
algorithms and literature. 

  A literature bank of financial sequential datasets with varying complexity 
was created. This provides a source to the solution of existing literature 
inconsistencies and deficiencies in explaining the performance of deep learning 
artefacts for modelling irregular sequential behaviour see Section~\ref{sec:annexures}.

  It is recommended to further develop the SeLFISA framework to create a 
consistent shareable technical and literature platform. The framework will 
consist of a knowledge bank of implemented irregular sequential analysis 
frameworks, datasets, algorithms and literature.

\section{Annexures}
\label{sec:annexures}
The annexures of results and datasets can be found on GitHub:
\ifpeerreview
\url{https://github.com/Anonymous/...}
\else
\url{https://github.com/Dandajena/SATNAC_2021_Paper}.
\fi

\section{Acknowledgment}
This work is funded by the 
\ifpeerreview
Anonymized.
\else
Research Committee of the University of the Western 
Cape and Telkom/Aria Technology Africa Centre of Excellence. The work received 
computational processing environment support from the South African Centre 
for High Performance Computing.
\fi
\bibliographystyle{IEEEtran}
\bibliography{2021_SATNAC}
\newpage              
\ifpeerreview
\else
\begin{IEEEbiographynophoto}{Kudakwashe Dandajena} 
is a Ph.D. student in Computer Science at the University of the Western 
Cape, investigating the optimisation of deep learning algorithms 
in discrete irregular sequential patterns.  He is program manager for the 
community of scientists of the Next Einstein Initiative at the African Institute 
for Mathematical Sciences Global Secretariat in Rwanda. Kuda holds a Master's
 degree in Computer Science.  Mobile: +25078~582~1565.
\end{IEEEbiographynophoto}
\vspace{-30pt}
\begin{IEEEbiographynophoto}{Isabella M. Venter}
is an Emeritus Professor in Computer Science at the University of the Western 
Cape, South Africa and the Chair of the Management Committee of 
the Telkom/Aria Technologies Africa Centre of Excellence. Her research interests 
include: computer science education, human computer interaction and information 
communication technologies for development.  Mobile: +2782~202~3520.
\end{IEEEbiographynophoto}
\vspace{-30pt}
\begin{IEEEbiographynophoto}{Mehrdad Ghaziasgar}
holds a PhD in Computer Science and is a Senior Lecturer in Computer Science 
at the University of the Western Cape. He heads up the Assistive Technologies 
Research Group at the Department which investigates applications of computer 
vision and machine learning to novel assistive technologies.  Tel: +2721~959 3012. 
\end{IEEEbiographynophoto}
\vspace{-30pt}
\begin{IEEEbiographynophoto}{Reg Dodds} has lectured in Computer Science 
at Stellenbosch, Natal, Port Elizabeth Universities and at the University 
of the Western Cape.
\end{IEEEbiographynophoto}
\fi

\vfill
\end{document}